\IfFileExists{./prepreamble-amspreprint.sty}{\RequirePackage[packages,theorems,changes]{./prepreamble-amspreprint}}{}

\makeatletter
\IfFileExists{./scoop-latex/scoop-packages.sty}{\providecommand*{\input@path}{}\edef\input@path{{./scoop-latex/}\input@path}}{}
\makeatother
\PassOptionsToPackage{style=authoryear-comp,backend=biber}{biblatex}
\PassOptionsToPackage{commentmarkup = footnote}{changes}    

\documentclass[english]{amsart}

\RequirePackage{biblatex}
\RequirePackage{ltxcmds}
\IfFileExists{./preamble-amspreprint.sty}{\RequirePackage[packages,theorems,changes]{./preamble-amspreprint}}{}

\usepackage{scoop-local}

\IfFileExists{./postpreamble-amspreprint.sty}{\RequirePackage[packages,theorems,changes]{./postpreamble-amspreprint}}{}

\makeatletter
\@ifpackageloaded{changes}{
\definechangesauthor[name = {Frederik Köhne}, color = {red!80!black}]{FK}
\definechangesauthor[name = {Anton Schiela}, color = {blue!80!black}]{AS}
}{}
\makeatother        

\makeatletter
\@ifpackageloaded{hyperref}{%
	\hypersetup{
		pdftitle = {An Exponential Averaging Process with Strong Convergence Properties},
		pdfauthor = {Frederik Köhne, Anton Schiela},
		pdfkeywords = {exponential moving averages, stochastic gradient descent, stochastic convergence analysis, random dynamical systems},
		pdfcreator = {Created using the Scoop Template Engine version 1.2.0.}
	}
}{
	\pdfinfo{
		/Title (An Exponential Averaging Process with Strong Convergence Properties)
		/Author (Frederik Köhne, Anton Schiela)
		/Subject ()
		/Keywords (exponential moving averages, stochastic gradient descent, stochastic convergence analysis, random dynamical systems)
		/Creator (Created using the Scoop Template Engine version 1.2.0.)
	}
}
\makeatother

\title[Exponential Averaging Process with Strong Convergence Properties]{An Exponential Averaging Process with Strong Convergence Properties}

\author{Frederik Köhne\orcidlink{0009-0008-6185-9675}}
\address[F. Köhne]{Department of Mathematics, University of Bayreuth, 95440 Bayreuth, Germany}
\email{frederik.koehne@uni-bayreuth.de}
\urladdr{https://num.math.uni-bayreuth.de/en/team/frederik-koehne/}

\author{Anton Schiela\orcidlink{0000-0002-6959-2951}}
\address[A. Schiela]{Department of Mathematics, University of Bayreuth, 95440 Bayreuth, Germany}
\email{anton.schiela@uni-bayreuth.de}
\urladdr{https://num.math.uni-bayreuth.de/en/team/anton-schiela/}

\thanks{This work was supported by DFG grant SCHI~1379/8--1 within the Priority Program SPP~2298 (Mathematical Foundations of Deep Learning), which is gratefully acknowledged.}

\date{\today}

\dedicatory{}

\begin{document}

\begin{abstract}
Averaging, or smoothing, is a fundamental approach to obtain stable, de-noised estimates from noisy observations.
In certain scenarios, observations made along trajectories of random dynamical systems are of particular interest.
One popular smoothing technique for such a scenario is exponential moving averaging (EMA), which assigns observations a weight that decreases exponentially in their age, thus giving younger observations a larger weight.
However, EMA fails to enjoy strong stochastic convergence properties, which stems from the fact that the weight assigned to the youngest observation is constant over time, preventing the noise in the averaged quantity from decreasing to zero.
In this work, we consider an adaptation to EMA, which we call $p$-EMA, where the weights assigned to the last observations decrease to zero at a subharmonic rate.
We provide stochastic convergence guarantees for this kind of averaging under mild assumptions on the autocorrelations of the underlying random dynamical system.
We further discuss the implications of our results for a recently introduced adaptive step size control for Stochastic Gradient Descent (SGD), which uses $p$-EMA for averaging noisy observations.

\end{abstract}

\keywords{exponential moving averages, stochastic gradient descent, stochastic convergence analysis, random dynamical systems}

\makeatletter
\ltx@ifpackageloaded{hyperref}{%
\subjclass[2020]{\href{https://mathscinet.ams.org/msc/msc2020.html?t=60F15}{60F15}, \href{https://mathscinet.ams.org/msc/msc2020.html?t=60G10}{60G10}, \href{https://mathscinet.ams.org/msc/msc2020.html?t=60J20}{60J20}, \href{https://mathscinet.ams.org/msc/msc2020.html?t=68T05}{68T05}, \href{https://mathscinet.ams.org/msc/msc2020.html?t=90C15}{90C15}}
}{%
\subjclass[2020]{60F15, 60G10, 60J20, 68T05, 90C15}
}
\makeatother

\maketitle

\section{Introduction}
\label{sec:introduction}
Suppose we wish to estimate a quantity $\tau$, but we only have access to a sequence of noisy observations $(\widetilde \tau_n)_{n \in \N}$.
One straightforward way to get an estimate $\widehat \tau$ for $\tau$ with knowledge of the first $n$ observations $\widetilde \tau_1, \dots, \widetilde \tau_n$ is to use the arithmetic mean, i.e.
\begin{equation}
\tauclass_n = \frac1n\sum_{k=1}^n\widetilde \tau_k.
\label{eq:arithmetic-mean}
\end{equation}
If the observations $\widetilde \tau_k$ are independent, identically distributed (iid) random variables with finite variance, this leads to almost sure convergence of $\tauclass_n$ to the mean $\E{}{\widetilde \tau_1}$ by the strong law of large numbers (see, e.g., \cite[Theorem 10.10.22]{Bogachev2007}).
Also for more general settings results like the Birkhoff ergodic theorem ensure almost sure convergence, if $\widetilde \tau_k$ are observations made along the trajectory of an ergodic dynamical system (see, c.f., \cite[Theorem 2.3]{Krengel2011} or \cite[Corollary 2.5.2]{HernandezLerma2003}).
The arithmetic mean \eqref{eq:arithmetic-mean} assigns the same weight to every observation $\widetilde \tau _k$.
From an information theoretical point of view, this is reasonable if all the observations $\widetilde \tau_k$ carry the same amount of information about the current target $\tau$.
This is the case if the observations are iid or drawn along the trajectory of a stationary (stochastic) process.
If however the target $\tau$ also changes over time, one would like to assign younger observations a larger weight compared to older observations, while still having the beneficial effects of averaging (i.e. noise reduction, almost sure convergence).
A common approach towards this is \emph{exponential (moving) averaging} (EMA), also referred to as \emph{exponential smoothing}.
Here, the averaged observation $\tauema_n$ is updated using the recursion
\begin{equation}
\tauema_{n+1} = \gamma \tauema_n + (1 - \gamma) \widetilde \tau_{n+1}
\label{eq:ema}
\end{equation}
with some factor $\gamma \in (0,1)$ and some initialization $\tauema_0$.
Usually, $\gamma$ is selected to be \emph{close} to 1.
This type of averaging dates back to at least \cite{Brown1956}, and is nowadays a tool often used for time series analysis and signal processing.
It can also be found in the context of (stochastic) optimization, e.g. in momentum methods or modern optimizers from the machine learning literature.
An often neglected weakness of EMA is the lack of convergence of $\tauema_n$ to the mean of the observations $\widetilde \tau_n$, unless the noise in these observations vanishes over time.
This problem is caused by the fact, that the last observation always has a constant weight $(1-\gamma)$, which does not vanish, and so the noise in $\tauema_n$ is reduced compared to the noise in $\widetilde \tau_n$, but only by a constant factor related to $(1 - \gamma)$. Thus, EMA might be a good choice as averaging method if the observations $\widetilde \tau_k$ asymptotically become deterministic, i.e. the noise in $\widetilde \tau_n$ vanishes with $n \to \infty$.
An appealing way to combine the virtues of both methods is to consider time-dependent factors $\gamma$ in \eqref{eq:ema}:
\begin{equation}
\widehat \tau_{n+1} = \gamma_n \widehat \tau_n + (1 - \gamma_n) \widetilde \tau_{n+1}.
\label{eq:ema-time-dep}
\end{equation}
Such adaptations to EMA can be found in the literature \cite{Taylor2004, Gardner2006} in the context of time series analysis.
Clearly, to overcome the problems of EMA regarding convergence, one needs $\gamma_n \to 1$.
One choice for such a sequence could be $\gamma_n = 1 - \frac{1}{(n+1)^p}$ for some $p \in (\frac12,1)$.
We will refer to the sequence $\widehat \tau_n$ generated by \eqref{eq:ema-time-dep} with $\gamma_n = 1 - \frac{1}{(n+1)^p}$ as $p$-EMA and denote it as $\taupema_n$.
For $p = 1$, the recursion \eqref{eq:ema-time-dep} yields the same estimate as the classical arithmetic mean (if the initialization of $p$-EMA chosen as $\taupema_0 = \widetilde \tau_1$):
\[
\tauclass_n = \taupema_n \qquad \text{ if $p = 1$}.
\]
However, for $p < 1$ it is easy to see that:
\begin{enumerate}[label=\arabic*.]
\item The weight of the last observation $\widetilde \tau_{n+1}$ in \eqref{eq:ema-time-dep} vanishes with $n \to \infty$, enabling the noise in $\taupema_n$ to vanish as well.
\item For fixed $n$, the weight of $\widetilde \tau_k$ in $\taupema_n$ monotonically increases with $k \le n$, giving younger observations a larger weight compared to older observations.
\end{enumerate}
Note that the above conditions do not contradict each other.
While in the first, $n$ varies, it is fixed in the second, where $k$ is variable.
\begin{figure}
\includegraphics[width=\textwidth]{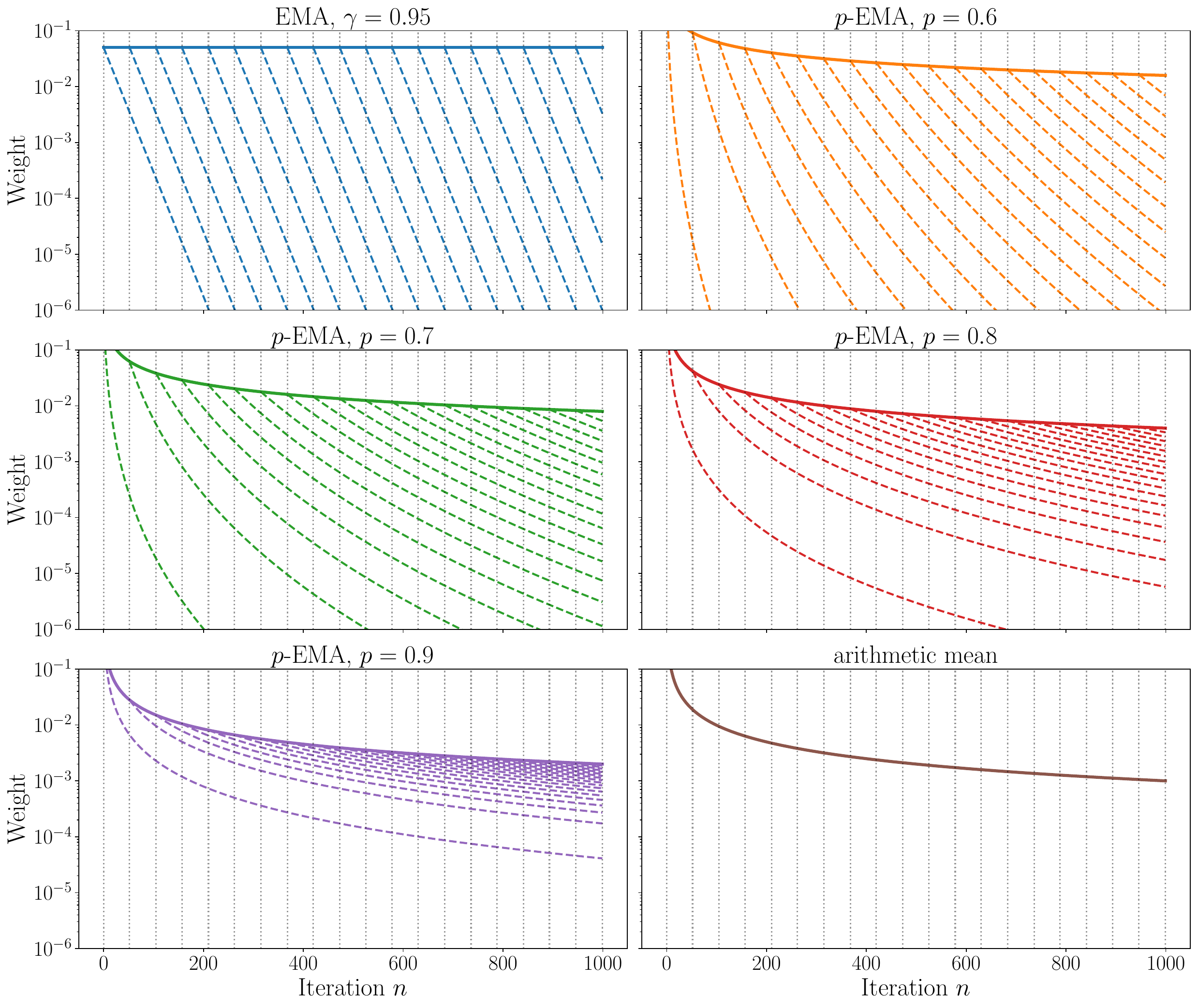}
\caption{Behavior of the weights in the different averaging procedures.}
\label{fig:weights}
\end{figure}
In each of the discussed averaging techniques (arithmetic mean, EMA, $p$-EMA), the estimate $\widehat \tau_n$ is a \emph{convex combination} of the observations $\widehat \tau_0, \widetilde \tau_1, \dots, \widetilde \tau_n$.
However, they differ in the distribution of weights,
which
is visualized in \cref{fig:weights}.
Each subplot shows the development of the weight assigned to the youngest observation $\widetilde \tau_n$ in $\widehat \tau_n$ (solid curve) in the corresponding averaging technique.
Additionally, each dashed line indicates the weight in $\widehat \tau_n$ assigned to the observations $\widetilde \tau_k$, where $k$ is the index, where the dashed line \emph{starts} (the indices $k$ we selected are indicated by the dotted vertical lines).
There are no dashed lines visible in case of the arithmetic mean, as all observations are assigned the same weight.
The solid curve is constant in case of EMA, as the youngest observation always has the constant weight $1 - \gamma$.
Qualitatively, we see that $p$-EMA yields an averaging technique \textit{between} EMA and the arithmetic mean.
In the present work, we will provide a rigorous stochastic convergence analysis for $p$-EMA, in particular for the case where the observations $\widetilde \tau_n$ are made along trajectories of sufficiently mixing random dynamical systems.
\\
Consequently, our results can be applied to any scenario, where noisy observations along the trajectory of such a (random) dynamical system are made.
As a particular application, we consider trajectories of the Stochastic Gradient Descent (SGD) algorithm with constant step sizes.
Utilizing that SGD can be interpreted as a random dynamical system, and that the trajectories of SGD converge to the support of a probability measure $\mu_\alpha^*$, invariant under the dynamics of SGD, we can show that for suitable observables $g: \Omega \times \R^d \to \R$, $\mu_\alpha^*$ almost every initial iterate $x_0$, and almost every realization of SGD starting at $x_0$, $p$-EMA applied to suitable observables $\widetilde \tau_k = g(\xi_k, x_k)$, converges to the mean of $g$ with respect to $P \times \mu_\alpha^*$:\footnote{The probability measure $P$ is introduced below.}
\[
\taupema_n \to \int_\Omega \int_{\R^d }g(\xi, x) \intd P(\xi)\intd\mu_\alpha^*(x),\, n \to\infty.
\]
This result has implications on the convergence theory of a recently developed adaptive step size scheme for SGD found in  \cite{KoehneKreisSchielaHerzog:2023:1}, where $p$-EMA is employed to smooth observations needed for adaptive step sizes for SGD.
\\
The rest of this paper is organized as follows:
In \cref{sec:convergence-p-ema} we provide the convergence analysis for $p$-EMA, using a technical construct we refer to as an  \emph{averaging scheme} and a generalized law of large numbers.
In \cref{app:sec:condition-on-p} we show that the restriction on $p$ for the factors $\gamma_n$ in \eqref{eq:ema-time-dep} is necessary to obtain almost sure convergence, even for independent observations.
In \cref{sec:sgd} we review SGD and its convergence to an invariant distribution, dependent on the step size.
The convergence results are then applied to the dynamics induced by SGD, with the special observables used for the adaptive step size estimation in \cref{sec:implications}.
Finally, we provide numerical results on $p$-EMA in general and applied to SGD trajectories in \cref{sec:numerics}.
\section{\texorpdfstring{Convergence of $p$-EMA}{Convergence of p-EMA}}
\label{sec:convergence-p-ema}
In this section we will provide the convergence analysis for $p$-EMA in a general form.
Consider a probability space $(\Gamma, \mathcal G, \pi)$ consisting of a set $\Gamma$, a $\sigma$-algebra $\mathcal G$ over $\Gamma$ and a measure $\pi : \mathcal G \to [0, 1]$ with $\pi(\Gamma) = 1$.
First, we introduce the notion of an \emph{averaging scheme}, show convergence results for this abstract class of weights and later show that $p$-EMA induces an averaging scheme in this sense.
From now on, we will drop the superscript $p$-EMA in $\taupema_n$, if it is evident from the context, that $\widehat \tau_n$ is obtained by $p$-EMA.
\subsection{Averaging Schemes}
We first consider weighted averages in general.
For a sequence $(b_n)_{n \in \N} \subset \R_{> 0}$ and a sequence of random variables $(X_n)_{n \in \N}$, we denote
\[
S_n = \sum_{k = 1}^n b_k X_k \qquad \text{ and } \qquad A_n = \sum_{k = 1}^n b_k.
\]
We are interested in convergence properties of the weighted average $\frac{S_n}{A_n}$.
For example, by the strong law of large numbers, one would expect convergence of $\frac{S_n}{A_n}$ to the mean $\E{}{X_1}$, if $b_k = 1$ and $(X_n)_{n \in \N}$ is a sequence of iid random variables with finite variance.
With the same selection of weights, ergodic theorems like the Birkhoff ergodic theorem (see, c.f., \cite[Theorem 2.3]{Krengel2011} or \cite[Corollary 2.5.2]{HernandezLerma2003}) provide convergence results, if $(X_n)_{n \in \N}$ are observations made along the trajectory of an ergodic random dynamical system.
In the following, we will derive a more general theory based on a recent result of \textcite{Korchevsky2010}, giving almost sure convergence, as in the strong law of large numbers or the Birkhoff ergodic theorem, but with general weighted averages, satisfying \cref{def:averaging-sequence} below.
\begin{definition}
By $\Psi_c$ we denote the set of all monotonically increasing functions $\psi: \R_{> 0} \to \R_{>0}$, such that
\begin{equation}
\sum_{n = 1}^\infty \frac{1}{n \psi(n)} < \infty.
\label{eq:def-avg-scheme}
\end{equation}
\end{definition}
For example, we have for $\eps > 0$ that $(x\mapsto x^\eps) \in \Psi_c$ and $(x \mapsto \log^{1+\eps}(x)) \in \Psi_c$.
\begin{definition}
\label{def:averaging-sequence}
A non-decreasing sequence $(b_n)_{n \in \N}$ of positive numbers is called \emph{averaging scheme}, if there is $\psi \in \Psi_c$, such that for all sufficiently large $n$ it holds:
\[
b_n\, \le  \frac{A_n}{\psi(A_n)}.
\]
\end{definition}
Intuitively this definition ensures that in the weighted average $\frac{1}{A_n}\sum_{k = 1}^n b_k \widetilde X_k$ the weight on the last observation is not too large compared to the previous weights.
\begin{remark}
The arithmetic mean induces an averaging scheme with $b_n = 1, A_n = n$ and $\psi (n) = n$.
EMA, however, does not induce an averaging scheme.
To see this, let us write EMA in the form
\[
\tauema = \frac{1}{A_n}\sum_{k = 1}^n b_k \widetilde \tau _k
\]
with some weights $b_n > 0$ and $A_n = \sum_{k = 1}^n b_k$.
It can easily be verified, that such a representation exists, if the initialization $\widehat \tau_0 = \widetilde \tau_1$ is chosen.
Then, we have
\begin{equation}
A_{n+1} = A_n + b_{n+1} \qquad \textnormal{implying} \qquad 1 = \frac{A_{n}}{A_{n+1}} + \frac{b_{n+1}}{A_{n+1}}.
\label{eq:remark:ema-no-avg-scheme:1}
\end{equation}
$\frac{b_{n+1}}{A_{n+1}}$ is the weight of the last observation $\widetilde \tau _{n+1}$ in $\widehat \tau_{n+1}$, and thus by definition of EMA equal to $(1 - \gamma)$.
Therefore,
\begin{equation}
\gamma = \frac{A_n}{A_{n+1}} \qquad \textnormal{ and thus } \qquad A_n = \gamma A_{n+1}.
\label{eq:remark:ema-no-avg-scheme:2}
\end{equation}
By inserting \eqref{eq:remark:ema-no-avg-scheme:2} into \eqref{eq:remark:ema-no-avg-scheme:1} we get:
\begin{equation}
b_{n+1} = (1-\gamma)A_{n+1}
\label{eq:remark:ema-no-avg-scheme:3}
\end{equation}
This contradicts the definition of an averaging scheme:
Suppose there is $\psi \in \Psi_c$ such that \eqref{eq:def-avg-scheme} is satisfied.
Then, necessarily $\psi(x) \to \infty, x \to \infty$.
Also, we have by \eqref{eq:remark:ema-no-avg-scheme:2} $A_n = \gamma^{-n+1}A_1$, and therefore $A_n \to \infty, n \to \infty$.
In particular, \eqref{eq:def-avg-scheme} would imply $\frac{b_n}{A_n} \to 0, n \to \infty$.
This contradicts \eqref{eq:remark:ema-no-avg-scheme:3}, which implies $\frac{b_n}{A_n} = (1 - \gamma) > 0$.
$p$-EMA on the other hand \textbf{does} induce an averaging scheme, as we will show below in \cref{subsec:pEMA-is-averaging-scheme}.
\end{remark}
Before we establish convergence results along trajectories under suitable dynamics, we focus on the more general case of dependent random variables.
Our result on averaging schemes, stated in \cref{thm:main-convergence-result} below, is a consequence of generalized law of large numbers found in \cite[Theorem 1]{Korchevsky2010}, which we state here without proof and refer the reader to the original work.
\begin{theorem}[{\cite[Theorem 1]{Korchevsky2010}}]
\label{thm:general-slln}
Consider a sequence of non-negative random variables $X_n$. Let $\{b_n\}$ be a sequence of positive numbers.
Suppose that the following conditions hold: $A_n \to \infty$ as $n \to \infty$,
\begin{equation}
\sum_{k=m}^n b_k \E{}{X_k} \leq C \sum_{k=m}^n b_k
\label{eq:general-slln:1}
\end{equation}
for all sufficiently large $n-m$, where $C$ is a constant, and
\begin{equation}
\E{}{\abs{S_n - \mathbb{E}S_n}^2} = O\left(\frac{A_n^2}{\psi(A_n)}\right)
\label{eq:general-slln:2}
\end{equation}
for some function $\psi \in \Psi_c$.
\bigskip
Then
\[
\frac{S_n - \mathbb{E}S_n}{A_n} \to 0 \quad \text{a.s.}
\]
\end{theorem}
As a consequence we obtain the following convergence result for averaging schemes in the sense of \cref{def:averaging-sequence}.
\begin{theorem}
\label{thm:main-convergence-result}
Consider a sequence of random variables $(X_n)_{n \in \N}$ on $(\Gamma, \mathcal G, \pi)$ such that:
\begin{enumerate}[label=\arabic*.]
\item $\E{}{X_n} = \E{}{X_1} =: \eta$ for all $n \in \N$.
\item\label{thm:main-convergence-result:item:autocorrelation-summability} $\abs{\E{}{X_nX_m} - \eta^2} \le \rho(\abs{n - m})$ for some function  $\rho : \N_0 \to \R_{\ge 0}$.
\item $\sum_{m = 0}^\infty \rho(m) < \infty$.
\item $X_n \ge c\; \as$ for some $c \in \R$ and all $n \in \N$.
\end{enumerate}
Further, suppose $(b_n)_{n \in \N}$ is an averaging scheme. Then:
\[
\frac{1}{A_n}\sum_{k = 1}^n b_k X_k \to \eta \qquad \pi-\text{almost surely}.
\]
\end{theorem}
\begin{proof}
By considering $X_n - c$ instead of $X$, we can assume that $X_n \ge 0$.
We seek to apply \cref{thm:general-slln}.
For this, we have to verify \cref{eq:general-slln:1,eq:general-slln:2}.
In our setting $\E{}{X_n} = \E{}{X_1}$ implies \eqref{eq:general-slln:1} with equality ($C = 1$).
Denote $\eta = \E{}{X_1}$.
For \eqref{eq:general-slln:2} we compute:
\begin{align*}
\E{}{\abs{S_n - \E{}{S_n}}^2}
&=
\E{}{\abs{S_n - A_n\eta}^2}
\\
&=
\sum_{k,\ell = 1}^n b_kb_\ell\left(\int X_kX_\ell d\mu - \eta^2\right)\\
&\le
2\sum_{m = 0}^{n-1} \rho(m)\sum_{i = 1}^mb_ib_{m-i}
\\
&\le
2\sum_{m = 0}^\infty \rho(m) b_nA_n
\\
&=
2\sum_{m = 0}^\infty \rho(m) \frac{b_n}{A_n}A_n^2
\\
&\lesssim
\frac{A_n^2}{\psi(A_n)},
\end{align*}
which verifies \eqref{eq:general-slln:2}.
Thus, by \cref{thm:general-slln}:
\[
\frac{S_n - \E{}{S_n}}{A_n} \to 0 \qquad \as
\]
On the other hand, we have $\E{}{S_n} = A_n \E{}{X_1}$ and therefore $ \frac{S_n - \E{}{S_n}}{A_n} = \frac{S_n}{A_n} - \E{}{X_1}$. Thus, $\frac{S_n}{A_n} \to \E{}{X_1}$ almost surely.
\end{proof}
\subsection{\texorpdfstring{$p$-EMA}{p-EMA} Induces an Averaging Scheme}
\label{subsec:pEMA-is-averaging-scheme}
In this subsection, we show that \cref{thm:main-convergence-result} can be applied to $p$-EMA.
Recall the definition of $p$-EMA. We select an initialization $\widehat \tau_0$ and update according to
\begin{equation}
\widehat \tau_{n+1} = \gamma_n \widehat \tau_n + (1 - \gamma_n) \widetilde \tau_{n+1}
\label{eq:recall-p-EMA}
\end{equation}
with $\gamma_n = 1 - \frac{1}{(n+1)^p}$.
We might choose $\widehat \tau_0 = \widetilde \tau_1$.\footnote{The concrete choice of initialization is irrelevant for convergence properties.}
Then, explicitly writing the recursion \eqref{eq:recall-p-EMA} we get:
\begin{align*}
\widehat \tau_{n+1} &= \left(1 - \frac{1}{(n+1)^p}\right)\widehat \tau_n + \frac{1}{(n+1)^p} \widetilde \tau_{n+1}
\\
&
=
\left(1 - \frac{1}{(n+1)^p}\right)\left(1 - \frac{1}{n^p}\right) \widehat \tau_{n-1} \\
&
\qquad
+\left(1 - \frac{1}{(n+1)^p}\right)\frac{1}{n^p}\widetilde \tau_{n} + \frac{1}{(n+1)^p}\widetilde \tau_{n+1}
\\
&
\;\;\vdots
\\
&
=
\widehat \tau_1\prod_{k = 2}^{n+1}\left(1 - \frac{1}{k^p}\right) + \sum_{k = 2}^{n+1}\widetilde \tau_k\frac{1}{k^p}\prod_{s = k+1}^{n+1}\left(1-\frac{1}{s^p}\right)
\\
&
=
\sum_{k = 1}^{n+1}\beta^{(n+1)}_k\widetilde \tau_k
\end{align*}
with
\[
\beta^{(n+1)}_k = k^{-p}\prod_{s = k+1}^{n+1}\left(1-\frac{1}{s^p}\right).
\]
A crucial step is to factor $\beta_k^{(n+1)}$ into a part depending only on $n+1$, and a part only depending on $k$.
By expanding the product we obtain:
\begin{equation}
\beta_k^{(n+1)} =
\left[
\prod_{s = 2}^{n+1}\left(1 - \frac{1}{s^p}\right)
\right]
\underbrace{
\left[
k^{-p}\prod_{s = 2}^k\left(1 - \frac{1}{s^p}\right)^{-1}
\right]
}
_{\beta_k \coloneqq}
.
\label{eq:def-beta}
\end{equation}
By construction, we have $\sum_{k = 1}^n \beta_k^{(n)} = 1$, and thus
\[
\Lambda_n \coloneqq \sum_{k = 1}^n \beta_k =\left[
\prod_{s = 2}^{n+1}\left(1 - \frac{1}{s^p}\right)
\right]^{-1}.
\]
We also have $\frac{\beta_n}{\Lambda_n} = n^{-p}$, as this is the weight of $\widetilde \tau_n$ in $\widehat \tau_n$.
Thus, a candidate for an averaging scheme is $(\beta_n)_{n \in \N}$, and for $p$-EMA we obtain
\[
\widehat \tau_n = \frac{1}{\Lambda_n} \sum_{k = 1}^n \beta_k \widetilde \tau_k.
\]
In this weighted average, there is, as desired, more weight on younger observations:
\begin{lemma}
\label{lem:beta-k-monoton}
For $p < 1$, the sequence $(\beta_n)_{n \in \N}$ is monotonically increasing.
\end{lemma}
\begin{proof}
Since $\beta_k \neq 0$ for all $k$, we can show $\frac{\beta_{k}}{\beta_{k+1}} < 1$ for all $k$. By \eqref{eq:def-beta} we have:
\[
\frac{\beta_{k}}{\beta_{k+1}} = \frac{k^{-p}}{(k+1)^{-p}}\left(1 - \frac{1}{(k+1)^p}\right) = \frac{(k+1)^p - 1}{k^p}
\]
The latter is $< 1$, if and only if
\[
(k+1)^p < k^p + 1,
\]
which is true for $p < 1$.
\end{proof}
We will use the following notation:
\begin{definition}
Let $\mathcal S$ be some set and $f, g : \mathcal S \to \R$ be two functions.
Then, we write
\[
f(s) \lesssim g(s),
\]
if there is a uniform constant $c$, independent of $s \in \mathcal S$, such that $f(s) \le c g(s)$ for all $s \in \mathcal S$.
We will use this notation mutatis mutandis for sequences.
\end{definition}
The following result, whose proof is surprisingly involved, shows that, for appropriate $p$, this is indeed an averaging scheme.
\begin{proposition}
\label{prop:pEMA-induces-avg-scheme}
For $p \in (\frac{1}{2}, 1]$, there is $\eps > 0$, such that for $n$ sufficiently large
\[
\beta_n \le \frac{\Lambda_n}{\log^{1+\eps}(\Lambda_n)}.
\]
In particular, $(\beta_n)_{n \in N}$ is an averaging scheme in the sense of \cref{def:averaging-sequence} with $\psi(x) = \log^{1 + \eps}(x)$.
\end{proposition}
We comment on the necessity of the condition $p \in (\frac12, 1]$ in \cref{app:sec:condition-on-p}.
\begin{proof}
Note that $\frac{\beta_n}{\Lambda_n} = n^{-p}$, as $\frac{\beta_n}{\Lambda_n}$ is the weight of $\widetilde \tau_n$ in $\widehat \tau_n$ obtained by $p$-EMA. We will show
\begin{equation}
\lim_{n \to \infty} \frac{\log(\Lambda_n)}{n^\frac{p}{1+\eps}} = 0,
\label{eq:proof-avg-sqn:1}
\end{equation}
for any $\eps \in \left(0, \frac{2p-1}{1- p}\right)$ if $p < 1$ and any $\eps > 0$ if $p = 1$, which implies the result, as $\Lambda_n \to \infty, n \to \infty$.
The proof for \eqref{eq:proof-avg-sqn:1} will be given in multiple lemmas and is structured as follows:
\begin{enumerate}
\item We derive a differentiable function $\widetilde \Lambda : \R_{>0} \to \R$, such that $\Lambda_n \lesssim c_a + \widetilde \Lambda(n)$ for some constant $c_a$ and $n$ sufficiently large (\cref{lem:bound-on-Sigma-n}).
\item We show that the limit in \eqref{eq:proof-avg-sqn:1} aggress with the limit
\[
\lim_{y \to \infty} \frac{1+\eps}{p}\frac{(y+1)^{-p}}{y^{\frac{p}{1+\eps} - 1}}\frac{g(y)}{c_a + \widetilde \Lambda(y)}
\]
for some function $g : \R_{>0} \to \R$.
We show that the second fraction converges to zero (\cref{lem:convergence:1}), while the third fraction converges to one (\cref{lem:convergence:2}).
\end{enumerate}
\begin{lemma}
\label{lem:bound-on-Sigma-n}
Define the mapping
\begin{align*}
\widetilde \Lambda : \R_{>0} &\to \R \\
y &\mapsto \int_2^{y+1} s^{-p} \exp\left(\int_2^{s+1} \log\left(\frac{\tau^p}{\tau^p - 1}\right)\intd\tau\right)\intd s.
\end{align*}
Then $\widetilde \Lambda$ is monotonically increasing and there is an additive constant $c_a$, such that
\[
\Lambda_n \lesssim c_a + \widetilde \Lambda(n).
\]
\end{lemma}
\begin{proof}[Proof of \cref{lem:bound-on-Sigma-n}]
Observe that the mapping $y \mapsto \frac{y^p}{y^p - 1}$ is monotonically decreasing in $y > 1$. Thus, the same holds for $\log\left( \frac{y^p}{y^p - 1}\right)$. In particular, we have
\[
\sum_{j = 2}^k \log\left( \frac{j^p}{j^p - 1}\right) \le \log\left(\frac{2^p}{2^p - 1}\right) + \int_2^k \log\left(\frac{\tau^p}{\tau^p -1}\right)\intd\tau.
\]
Thus, we derive:
\[
\prod_{j = 2}^k \frac{j^p}{j^p-1} = \exp\left(\sum_{j = 2}^{k}\log\left(\frac{j^p}{j^p - 1}\right)\right) \lesssim \exp\left(\int_2^k \log\left(\frac{\tau^p}{\tau^p - 1}\right)\intd\tau\right)
\]
Now consider the function
\[
h(y) \coloneqq y^{-p} \exp\left(\int_2^y \log\left(\frac{\tau^p}{\tau^p - 1}\right)\intd\tau\right).
\]
We have
\[
h'(y) = y^{-p}\left(-py^{-1} + \log\left(\frac{y^p}{y^p - 1}\right)\right)\exp\left(\int_2^y \log\left(\frac{\tau^p}{\tau^p - 1}\right)\intd\tau\right).
\]
The term in the first bracket can be bounded as follows, using the well known inequality $\log(1 + x) \ge \frac{x}{1 + x}$ and $\frac{y^p}{y^p-1} = 1 + \frac{1}{y^p - 1}$:
\[
-p\inv y + \log\left(1 + \frac{1}{y^p - 1}\right) \ge -p\inv y + \frac{1}{y^p - 1} \frac{1}{1 + \frac{1}{y^p - 1}} = -p\inv y + y^{-p}
\]
and thus $h'(y) \ge 0$ for $y$ sufficiently large.
Further, we have
\[
\beta_k = k^{-p}\prod_{s = 2}^k\left(1 - \frac{1}{s^p}\right)^{-1} = k^{-p}\prod_{s = 2}^k \frac{s^p}{s^p - 1} \lesssim h(k).
\]
Therefore, there is a constant $c_a$, such that:
\[
\Lambda_n \lesssim \sum_{k = 1}^n h(k) \le c_a + \int_2^{n+1}h(y)\intd y = c_a + \widetilde \Lambda(n)
\]
\end{proof}
Next, we state one additional technical lemma.
\begin{lemma}
\label{lem:convergence:1}
It holds:
\[
\lim_{y \to \infty} \frac{(y + 1)^{-p}}{y^{\frac{p}{1+\eps} - 1}} = 0
\]
\end{lemma}
\begin{proof}[Proof of \cref{lem:convergence:1}]
This is trivial if $p = 1$.
Otherwise, the claimed convergence is equivalent to the convergence of
\[
\frac{y^{-p}}{y^{\frac{p}{1+\eps} -1}} = y^{-p(1 + \frac{1}{1+\eps}) + 1}
\]
and thus to $-p(1 + \frac{1}{1+\eps}) + 1 < 0$.
We compute:
\begin{alignat*}{3}
&\qquad&&\begin{array}[t]{l}-p\left(1+\frac{1}{1+\eps}\right) + 1\end{array} &&<\, 0 \\
&\iff\qquad &&\begin{array}[t]{l} p(2+\eps)\end{array}&&>\, 1+\eps\\
&\iff &&\begin{array}[t]{l} \eps(1-p)\end{array}&&<\, 2p-1
\end{alignat*}
By assumption we have $0 < \eps < \frac{2p - 1}{1 - p}$, such that the last assertion, and thus the claim, is true.
\end{proof}
\begin{lemma}
\label{lem:convergence:2}
Define
\[
g(y) = \exp\left(\int_2^{y+1}\log\left(\frac{\tau^p}{\tau^p - 1}\right)\intd\tau\right),
\]
such that $\widetilde \Lambda(y) = \int_2^{y+1} s^{-p}g(s)\intd s$. Then:
\begin{equation}
\lim_{y \to \infty} \frac{g(y)}{\widetilde \Lambda(y)} = 1
\label{eq:lemma-limit}
\end{equation}
\end{lemma}
\begin{proof}[Proof of \cref{lem:convergence:2}]
We have $\log(\frac{y^p}{y^p - 1}) = \log\left(1 + \frac{1}{y^p - 1}\right)$. Therefore:
\[
\frac{1}{y^p} = \frac{1}{y^p - 1}\frac{1}{1 + \frac{1}{y^p - 1}}\le\log\left(\frac{y^p}{y^p - 1}\right) \le \frac{1}{y^p - 1}
\]
Using this, we see that:
\[
g(y) \gtrsim \exp\left(\int_2^{y+1} \tau^{-p}\intd\tau\right) \to \infty, y \to \infty.
\]
Trivially, we have $\lim_{y \to \infty} \widetilde \Lambda(y) = \infty$ as well, so that
we will consider
\begin{equation}
\lim_{y \to \infty} \frac{g'(y)}{{\widetilde \Lambda}^\prime(y)}
\label{eq:alternative-limit}
\end{equation}
and the limits in \eqref{eq:lemma-limit} and \eqref{eq:alternative-limit} agree by L'Hôpital's rule.
We have
\[
g'(y) = \log\left(\frac{(y+1)^p}{(y+1)^p - 1}\right)g(y)
\]
and
\[
\widetilde \Lambda'(y) = (y+1)^{-p}g(y+1).
\]
It holds $g(y+1) - g(y) \to 0, y \to \infty$, and therefore $\frac{g(y)}{g(y+1)} \to 1, y \to \infty$, as $g(y) \to \infty, y \to \infty$.
It is not hard to show that
\[
\lim_{y \to \infty} y \log\left(1 + \frac{1}{y}\right) = 1,
\]
therefore:
\[
\frac{\log\left(\frac{(y+1)^p}{(y+1)^p - 1}\right)}{(y+1)^{-p}} \to 1, y \to \infty.
\]
Thus, we have
\[
\frac{g'(y)}{\widetilde \Lambda'(y)} = \frac{\log\left(\frac{(y+1)^p}{(y+1)^p - 1}\right)}{(y+1)^{-p}}\frac{g(y)}{g(y+1)} \to 1, y \to \infty,
\]
as both factors converge to $1$. This concludes the proof of \cref{lem:convergence:2}.
\end{proof}
Define $\widehat \Lambda(y) = c_a + \widetilde \Lambda(y)$. Then, by \cref{lem:bound-on-Sigma-n} we have for some constant $c_m$ (due to $\lesssim$ in the results above)
\begin{equation}
0 \le \frac{\log(\Lambda_n)}{n^{\frac{p}{1+\eps}}} \le \frac{\log(c_m) + \log(\widehat\Lambda(n))}{n^{\frac{p}{1+\eps}}}
.
\label{eq:proof-avg-sqn:2}
\end{equation}
We have $\log(\widehat \Lambda(y)) \to \infty, y \to \infty$ and $y^{\frac{p}{1+\eps}} \to \infty, y \to \infty$. Thus, the limit on the right-hand side of \eqref{eq:proof-avg-sqn:2} exists if the limit
\[
\lim_{y\to\infty} \frac{\widehat \Lambda '(y)}{\widehat \Lambda(y) \frac{p}{1+\eps} y^{\frac{p}{1+\eps} - 1}}
\]
exists, and in this case, the limits agree, again, by L'Hôpital's rule. Observe that
\[
\widehat \Lambda'(y) = \widetilde \Lambda'(y) = (y+1)^{-p} g(y).
\]
Thus:
\[
\frac{\widehat \Lambda '(y)}{\widehat \Lambda(y) \frac{p}{1+\eps} y^{\frac{p}{1+\eps} - 1}} = \frac{1+\eps}{p}\frac{(y+1)^{-p}}{y^{\frac{p}{1+\eps} - 1}}\frac{g(y)}{c_a + \widetilde \Lambda(y)}
\]
The first fraction is just a constant. For $y \to \infty$, the second converges to zero due to \cref{lem:convergence:1}, and the third converges to one due to \cref{lem:convergence:2} (as $c_a$ is a constant and $g(y), \widetilde \Lambda(y) \to \infty$). Thus, we conclude from \eqref{eq:proof-avg-sqn:2}:
\[
\lim_{n \to \infty} \frac{\log(\Lambda_n)}{n^{\frac{p}{1+\eps}}} = 0,
\]
which concludes the proof of \cref{prop:pEMA-induces-avg-scheme}.
\end{proof}
\subsection{Autocorrelations}
We seek to apply \cref{thm:main-convergence-result,prop:pEMA-induces-avg-scheme} to the case where the sequence of random variables $(X_n)_{n \in \N}$ represents a sequence of observations made along the trajectory of some (random) dynamical system.
In this scenario, we have a \emph{measure preserving} map of the probability space $(\Gamma, \cG, \pi)$ into itself, \ie\ a measurable function
\begin{equation}
\theta : \Gamma \to \Gamma
\qquad
\text{with}
\qquad
\pi(G) = \pi(\theta(G))
\text{ for all }
G \in \cG,
\label{eq:measure-preserving}
\end{equation}
and an \emph{observable} $g \in L_2(\Gamma)$ and consider $X_n(\omega) = g(\theta^n \omega)$, which is a sequence of (in general dependent) random variables on $(\Gamma, \cG, \pi)$.
Here, write $\theta \omega \coloneqq \theta (\omega)$ for $\omega \in \Gamma$.
Condition \ref{thm:main-convergence-result:item:autocorrelation-summability} in \cref{thm:main-convergence-result} now translates to a condition on the decay of autocorrelations of $\theta$ under the \emph{observable} $g$:
A direct consequence of \eqref{eq:measure-preserving} is:
\begin{equation}
\int_\Gamma g(\theta \omega)\intd\pi(\omega) = \int_\Gamma g(\omega)\intd\pi(\omega),
\label{eq:change-of-variable-measure-preserving}
\end{equation}
and thus:
\[
\int_\Gamma g(\theta^n\omega)g(\theta^m\omega)\intd\pi(\omega) = \int_\Gamma g(\omega)g(\theta^{\abs{m - n}}\omega)\intd\pi(\omega).
\]
This motivates the following definition:
\begin{definition}
\label{def:summable-decay-of-corrletations}
Consider $g \in L_2(\Gamma)$ and a measure preserving mapping $\theta : \Gamma \to \Gamma$.
\begin{enumerate}[label=\arabic*.]
\item For $m \in \N_0$, we define the \emph{coefficient of autocorrelation} as:
\[
\rho(m) \coloneqq \int_\Gamma g(\omega)g(\theta^m\omega)\intd \pi(\omega) - \left(\int_\Gamma g(\omega)\intd\pi(\omega)\right)^2
\]
\item We say $g$ has summable decay of correlations under $\theta$, if
\begin{equation}
\sum_{m = 0}^\infty \abs{\rho(m)} < \infty.
\label{eq:summability-assumption}
\end{equation}
\end{enumerate}
\end{definition}
If $g, g \circ \theta, g \circ \theta^2, \dots$ were independent random variables with finite variance, we would have $\rho(m) = 0$ for all $m > 0$, rendering the summability assumption \eqref{eq:summability-assumption} trivial.
In this sense, the assumption of summable decay of correlations quantifies the rate of mixing of $\theta$.
With this, we are ready to state the main result regarding convergence of $p$-EMA along trajectories:
\begin{theorem}
\label{thm:convergence-pEMA-trajectories}
Consider $p$-EMA with $p \in (\frac12,1]$, applied to observations $\widetilde \tau_n = g(\theta ^{n-1} \omega_0)$, where $g$ has summable decay of correlations under $\theta$ and is bounded from below. Then we have:
\[
\widehat \tau_n \to \int_\Gamma g \intd\pi
\]
for $\pi$-almost every $\omega_0$.
\end{theorem}
\begin{proof}
We can view the observations $\widetilde \tau_n = g \circ \theta^{n-1}$ as random variables on $\Gamma$. Since $\theta$ is measure preserving, we have (see \cref{eq:change-of-variable-measure-preserving}):
\[
\E{}{\widetilde \tau_n} = \int_\Gamma g(\theta^{n-1}\omega)\intd \pi(\omega) = \int_\Gamma g(\omega)\intd\pi(\omega) = \E{}{\widetilde \tau_1} =: \eta .
\]
Further, summable decay of correlations implies $\E{}{\widetilde \tau_n \widetilde \tau_m} - \eta^2 = \rho(\abs{n - m}) $ for some $\rho: \N_0 \to \R$ with
\[
\sum_{m = 0}^\infty\abs{\rho(m)} < \infty.
\]
Also, as $g$ is assumed to be bounded from below, all $\widetilde \tau_n$ are bounded from below.
Finally, we have for the average $\widehat \tau_n$, obtained by $p$-EMA:
\[
\widehat \tau_n = \frac{1}{\Lambda_n}\sum_{k = 1}^n \beta_k \widehat \tau_k,
\]
where $\Lambda_n = \sum_{k = 1}^n \beta_k$.
Recalling that $(\beta_n)_{n \in \N}$ is an averaging scheme (see \cref{prop:pEMA-induces-avg-scheme}), \cref{thm:convergence-pEMA-trajectories} follows from \cref{thm:main-convergence-result}.
\end{proof}
\section{On the Condition
\texorpdfstring{$p \in (\frac12,1]$}{p}
}
\label{app:sec:condition-on-p}
\begin{figure}
\includegraphics[width=\textwidth]{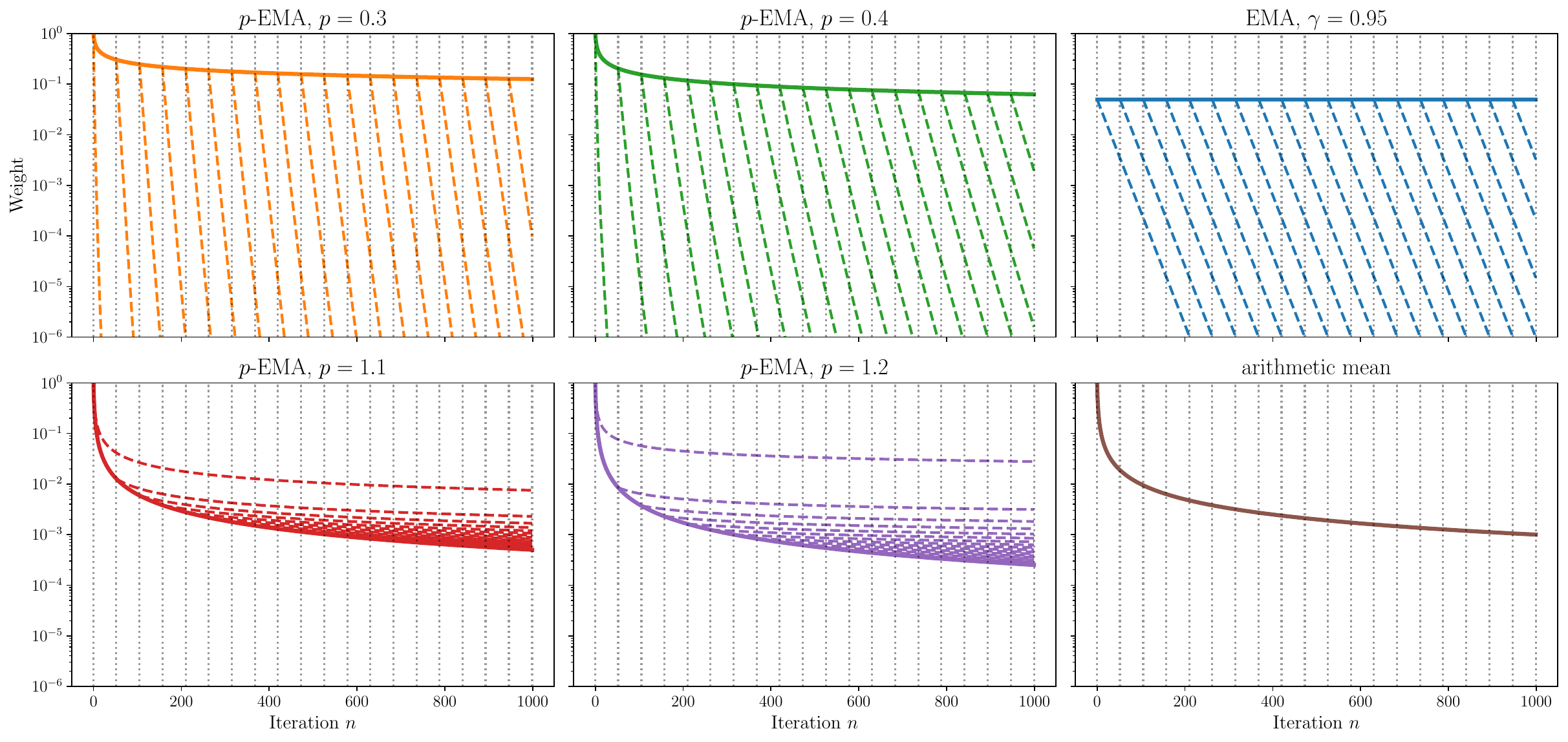}
\caption{Comparison of weights for $p$-EMA with $p$ outside the admissible interval $(\frac12, 1]$.}
\label{fig:bad-weights}
\end{figure}

We have imposed the condition $p \in (\frac12, 1]$, to show that $p$-EMA induces an averaging scheme in the sense of \cref{def:averaging-sequence}, and thus obtain convergence from \cref{thm:main-convergence-result}.
In fact, in the case $p = 1$ we have $\beta_n \equiv 1$ and $\Lambda_n = n$. Thus,
\[
\tau_n = \frac{1}{n}\sum_{k = 1}^n g(\theta^{k-1}\omega_0).
\]
Here, almost sure convergence is already known from the classical Birkhoff ergodic theorem in the more general case of an ergodic dynamical system $\theta$.
The assumption of summable decay of correlations (\cref{def:summable-decay-of-corrletations}) is a quantification of mixing and thus implies ergodicity.
The behavior of the weights of $p$-EMA with $p$ outside the interval $(\frac12, 1]$ is depicted in \cref{fig:bad-weights} (see also \cref{fig:weights} and its description in \cref{sec:introduction}).
In the case $p > 1$, we observe that older observations are assigned a \emph{larger} weight compared to younger observations.
Clearly, this is completely counterintuitive.
It also implies that, at any iteration, the sum of weights assigned to \emph{all} subsequent observations will stay uniformly bounded, a fact which we will use in \cref{subsec:case-p-greater-1} to show that we don't have almost sure convergence anymore, even on iid observations.
In the case $p < \frac12$, it's not that clear to see, want prevents almost sure convergence.
Loosely speaking, the weights assigned to younger observations don't decay fast enough, to ensure that the noise induced by sufficiently regularly occurring outliers is averaged out appropriately.
We will give a formal counterexample in \cref{subsec:case-p-smaller-12}, where almost sure convergence does not happen on iid observations.
The case $p = \frac12$ remains unclear and is subject to further research.
In this case, the weights assigned to the last observation are not square summable, a property that is evident for $p > 1/2$.
We believe that this property is crucial for ensuring almost sure convergence of the averaging technique.
However, the counterexample provided in \cref{subsec:case-p-smaller-12} does not apply to this case.
\subsection{The case
\texorpdfstring{$p > 1$}{p > 1}
}
\label{subsec:case-p-greater-1}
If $p > 1$, almost sure convergence can no longer be expected, even if all observations are iid. To see this, first observe that for any bounded sequence of observations $\widetilde \tau_n$, the sequences of differences $\abs{\tau_{n+1} - \tau_n}$ is summable:
\begin{align*}
\abs{\tau_{n+1} - \tau_n}
&
=
\frac{\abs{\Lambda_{n} S_{n+1} - \Lambda_{n+1} S_n}}{\Lambda_n\Lambda_{n+1}}
\\
&
=
\frac{\abs{\Lambda_{n} (S_{n} + \beta_{n+1}\widetilde \tau_n) - (\Lambda_{n} + \beta_{n+1})S_n}}{\Lambda_n\Lambda_{n+1}}
\\
&
\le
\frac{\beta_n\abs{\widetilde \tau_n}}{\Lambda_n} + \frac{\beta_{n+1}}{\Lambda_{n+1}}\frac{\abs{S_n}}{\Lambda_n}
\end{align*}
If $(\widetilde \tau_n)$ is a bounded sequence, so is $\frac{S_n}{\Lambda_n}$. The identity $\frac{\beta_n}{\Lambda_n} = n^{-p}$ implies that $\abs{\tau_{n+1} - \tau_n}$ is summable if $p > 1$. A concrete counterexample, where we do not have almost sure convergence now can be constructed as follows. Choose $N_0$, such that
\[
\sum_{n = N_0+1}^\infty n^{-p} < \frac14.
\]
Consider a sequence of iid random variables $X_n$, such that $P(X_n = 1) = P(X_n = -1) = \frac12$. Then, the event
$A = \{X_1 = \cdots = X_{N_0} = 1\}$ has probability $2^{-N_0} > 0$. However, on $A$ we do not have convergence of $\frac{S_n}{\Lambda_n}$ to $\E{}{X_1} = 0$:
\[
\frac{S_n}{\Lambda_n} = \tau_{N_0} + \tau_n - \tau_{N_0} \ge 1 - 2\sum_{n = N_0 + 1}^\infty n^{-p} > \frac12 \qquad \forall n > N_0.
\]
\subsection{The case
\texorpdfstring{$p < \frac12$}{p < 1/2}
}
\label{subsec:case-p-smaller-12}
For $p < \frac{1}{2}$, there is $s > 3$, such that $p(1-s) > -1$. Consider a sequence of random variables $(X_n)_{n \in \N}$ on $[1,\infty)$, independently and identically distributed according to the density function
\[
f(x) = \frac{1}{I_s}x^{-s},
\]
denoting $I_s = \int_1^\infty x^{-s}\intd x$. As $s > 3$, these random variables have finite first and second moment. In particular, they satisfy the assumptions of \cref{thm:main-convergence-result}. Then we have:
\[
P(X_n \ge 2n^p) = \frac{1}{I_s}\int_{2n^p}^\infty x^{-s} \intd x = \frac{1}{I_s}\frac{2^{p(1-s)}}{s - 1}n^{p(1 - s)}
\]
From $p(1 - s) > -1$ we conclude
\[
\sum_{n = 1}^\infty P(X_n \ge 2n^p) = \frac{2^{p(1-s)}}{I_s}\frac{1}{s - 1} \sum_{n = 1}^\infty n^{p(1 - s)} = \infty.
\]
All the events $A_n \coloneqq \{X_n \ge 2n^p\}$ are independent, thus, by the second Borel-Cantelli lemma, infinitely many of them occur with probability one.
We further have $\eta = \E{}{X_n} = \E{}{X_1} = \frac{1}{I_s} \int_1^\infty x^{1-s}\intd x = \frac{s-1}{s-2} = 1 + \frac{1}{s-2} < 2$.
However, on $A_n$ we have for the estimate $\widehat \tau_n$ obtained by $p$-EMA with observations $X_n$:
\[
\widehat \tau_n = \gamma_{n-1}\widehat \tau_{n-1} + (1 - \gamma_n) X_n \ge 1 - \frac{1}{n^p} + \frac{1}{n^p} X_n \ge 1 - \frac{1}{n^p} + \frac{1}{n^p} 2n^p = 3 - \frac{1}{n^p}
\]
Here, we have used that $\widehat \tau_{n-1} \ge 1$, as all observations are $\ge 1$, and that $X_n \ge 2n^p$ on $A_n$.
Thus, $\widehat \tau_n$ will escape any sufficiently small $\eps$-Ball around $\eta < 2$ with probability one infinitely often, contradicting almost sure convergence.
\section{SGD and Invariant Measures}
As a special case of application, we consider the Stochastic Gradient Descent method.
Its trajectories can be understood as the trajectories of a random dynamical system.
Hence, our results on averaging by $p$-EMA can be applied to observables evaluated along such trajectories.
This becomes of interest when such evaluations are used to determine step sizes online.
We will elaborate this in more detail in \cref{sec:implications}, and provide the relevant background in this section.
\label{sec:sgd}
\subsection{Stochastic Gradient Descent}
Consider a probability space $(\Omega, \mathcal A, P)$.
Suppose the function $f : \Omega \times \R^d \to \R$ satisfies:
\begin{enumerate}[label=\arabic*.]
\item For all $\xi \in \Omega$, the mapping x $\mapsto f(\xi, x)$ is convex and $L_\xi$-smooth for a measurable map $\xi \mapsto L_\xi$ satisfying
\[
L \coloneqq \esssup_{\xi} L_\xi < \infty.
\]
\item For all $x \in \R^d$, the mappings $\xi \mapsto f(\xi, x)$ and $\xi \mapsto \nabla_x f(\xi, x)$ are measurable and square-integrable.
\end{enumerate}
In this scenario, the mean
\[
F(x) = \int_\Omega f(\xi, x) \intd P(\xi)
\]
is also differentiable with
\[
\nabla F(x) = \int_\Omega \nabla_xf(\xi, x)\intd P(\xi).
\]
For simplicity of notation we will use $f_\xi = f(\xi, \cdot)$ and $\nabla f_\xi = \nabla _x f(\xi, \cdot)$.
We will assume that there is a measurable set $\widetilde \Omega \subset \Omega$ such that $f_{\xi}$ is $\mu_\xi$-strongly convex for some $\mu_\xi > 0$ for all $\xi \in \widetilde \Omega$ and $P(\widetilde \Omega) > 0$.
For example, in the finite sum setting, this is fulfilled, if at least one sampled functions is strongly convex.
Stochastic Gradient Descent (SGD) (first introduced by \cite{RobbinsMonro:1951:1}) with step size $\alpha$ can be given as the iteration
\begin{equation}
x_{k+1} = \varphi_\alpha(\xi_k, x_k),
\label{eq:SGD-iteration}
\end{equation}
where $\varphi_\alpha(\xi, x) = x - \alpha \nabla f_\xi(x)$ and $\xi_k \sim P$ is chosen randomly at each iteration.
This model on the noise in the search direction captures most practical scenarios such as single sample or mini-batch SGD.
\subsection{Interpolating vs. Non-Interpolating Setting}
\label{subsec:interpolating-vs-non-interpolating}
Two scenarios are distinguished in the convergence theory for SGD.
In the so-called interpolating setting, the noise in the search directions vanishes at the minimizer $x^*$. This means that we have
\begin{equation}
\E{\xi}{\norm{\nabla f_\xi(x^*)}^2} = 0.
\label{eq:interpolating-condition}
\end{equation}
The name stems from the fact that, for machine learning models, this is the case in the heavily overparameterized case, where the model is able to interpolate the training data. In the non-interpolating setting, the expectation in \eqref{eq:interpolating-condition} is positive.
In the convergence theory of SGD it is well known that, in the non-interpolating setting, the step sizes of SGD need to decrease to zero in order to ensure convergence to the minimizer.
\subsection{Invariant Measures}
In recent years, the stationary distribution of the iterates of SGD has gathered the interest of researchers \cite{Dieuleveut2020,Azizian2024,Shirokoff2024}.
Formally, a stationary distribution is a probability measure $\mu_\alpha^*$ on the state space $\R^d$ of SGD iterates, which is \emph{invariant} under the Markov Process induced by \eqref{eq:SGD-iteration}. For a Borel set $B \in \R^d$, denote the probability for $x_1$ belonging to $B$, given $x_0$, by $P(B, x_0)$, i.e.:
\[
P(B, x_0) = P(\xi \mid \varphi_\alpha(\xi, x_0) \in B) = P(\varphi(\cdot, x_0)^{-1}(B))
\]
Then, $\mu_\alpha^*$ satisfies
\begin{equation}
\mu_\alpha^*(B) = \int_\Omega P(B, x) \intd \mu_\alpha^*(x)
\label{eq:invariant-measure}
\end{equation}
for every Borel set $B \subset \R^n$.
More intuitively, this means that we have the implication (see also \cite{Azizian2024})
\[
x_0 \sim \mu_\alpha^* \, \Longrightarrow \, x_1 \sim \mu_\alpha^*.
\]
Existence and uniqueness of such an invariant measure have been discussed recently in different works, borrowing techniques from the theory of (random) dynamical systems and Markov processes.
Under our assumptions, we have:
\begin{theorem}
\label{thm:existence-uniqueness-ivpm}
For sufficiently small $\alpha$, there is a unique probability measure $\mu_\alpha^*$ which satisfies \eqref{eq:invariant-measure}.
\end{theorem}
\begin{proof}
We will use a well known result, namely that Markov Chains, whose transition functions are contracting on average, exhibit a unique invariant measure.
Such a result can be found \eg\ in \cite[Theorem 4.31]{Benaim2022}, and requires:
\begin{enumerate}[label=\arabic*.]
\item \begin{equation}
\int_\Omega \log(\ell_\xi) \intd P(\xi) =: -c < 0
\label{eq:contracting-on-avg}
\end{equation}
with some $c > 0$.
Here, $\ell_\xi$ is a Lipschitz constant for the mapping $x \mapsto \varphi_\alpha(\xi, x)$.
\item
\begin{equation}
\int_\Omega \max\left(\log \left(\norm{\varphi_\alpha(\xi, x_0) - x_0}\right), 0\right) \intd P(\xi) < \infty
\label{eq:fix-point-contraction}
\end{equation}
for some $x_0 \in \R^d$.
\end{enumerate}
\noindent
In our case, we consider the Markov Chain, generated by
\[
x_{k+1} = \varphi_\alpha(\xi_k, x_k),
\]
see \cref{eq:SGD-iteration}.
For $\xi \in \Omega$ and $x,y \in \R^d$, we infer:
\begin{align*}
\norm{\varphi_\alpha (\xi, x) - \varphi_\alpha (\xi, y)}^2
&=
\norm{x - y}^2 - 2\alpha \inner{x - y}{\nabla f_\xi(x) - \nabla f_\xi(y)}\\
&
\qquad
+ \alpha^2 \norm{\nabla f_\xi(x) - \nabla f_\xi(y)}^2\\
&\le
\norm{x - y}^2 + \alpha(\alpha L_\xi - 2) \inner{x - y}{\nabla f_\xi(x) - \nabla f_\xi(y)},
\end{align*}
using co-coercivity of $f_\xi$, which follows from convexity and $L_\xi$-smoothness (see \cite[Lemma 2.29]{GarrigosGower:2023:1}).
For $\alpha \le \frac{1}{L} \le \frac{1}{L_\xi}$, we have $\alpha L_\xi - 2 < -1$.
Noting that $\inner{x - y}{\nabla f_\xi(x) - \nabla f_\xi(y)} \ge \mu_\xi\norm{x - y}^2$ due to convexity, we have for sufficiently small $\alpha$
\[
\norm{\varphi_\alpha (\xi, x) - \varphi_\alpha (\xi, y)}^2 \le (1 - \alpha \mu_\xi)\norm{x - y}^2
\]
and thus the mapping $x \mapsto \varphi_\alpha(\xi, x)$ has a Lipschitz constant $\ell_\xi \le \sqrt{1 - \alpha \mu_\xi}$.
Consequently, we have
\[
\int_\Omega \log (\ell_\xi) \intd P(\xi) =: - c < 0
\]
for some $c > 0$, as we have assumed that $\mu_\xi > 0$ on the set of positive measure $\widetilde \Omega$.
This shows \eqref{eq:contracting-on-avg}.
To see that \eqref{eq:fix-point-contraction} holds as well, observe that for any $x_0$ we have
\[
\varphi_\alpha(\xi, x_0) - x_0 = -\alpha \nabla f_\xi(x_0).
\]
Thus we have:
\[
\int_\Omega \max\left(\log \left(\norm{\varphi_\alpha(\xi, x_0) - x_0}\right), 0\right) \intd P(\xi) \le \alpha \int_\Omega \norm{\nabla f_\xi(x_0)}\intd P(\xi) < \infty,
\]
as we have assumed that the map $\xi \mapsto \nabla f_\xi(x_0)$ is square-integrable (which implies integrability).
\end{proof}
This invariant measure from \cref{thm:existence-uniqueness-ivpm} satisfies:
\[
\int_{\R^d} g(x)\intd\mu_\alpha^*(x) = \int_{\R^d} \int_\Omega g(\varphi_\alpha(\xi, x)) \intd P(\xi)\intd \mu_\alpha^*(x)
\]
for any integrable $g$.
\section{Implications on Adaptive Step Size Estimators}
\label{sec:implications}
In this section, we consider a concrete example, where samples are made along a trajectory which eventually becomes stationary.
As it is demonstrated above, under certain assumptions, SGD with constant step sizes $\alpha$ exhibits a unique invariant probability measure $\mu_\alpha^*$.
If the algorithm is not started within the support of this measure, its iterates either diverge or converge towards the support of this measure.
Global convergence results can guarantee that the former does not happen, so we might assume that the iterates of SGD eventually become stationary, distributed according to the invariant measure $\mu_\alpha^*$.
This is a scenario, where $p$-EMA might be advantageous compared against
\begin{itemize}
\item the classical arithmetic mean, because the influence of early observations made when the trajectory was not yet distributed according to $\mu_\alpha^*$ decays faster.
\item classical EMA, because there still is noise in the observations, as the iterates of SGD move within the support of the invariant measure.
\end{itemize}
Of particular interest in application are the quantities
\begin{align*}
g_k = \E{\omega} {\norm{f_\omega'(x_k)}^2}
\qquad
\text{ and }
\qquad
\sigma_k &= \E{\omega} {\norm{f_\omega'(x_k) - F'(x_k)}^2}
\\
&= \E{\omega} {\norm{f_\omega'(x_k)}^2} - \norm{F'(x_k)}^2.
\end{align*}
As these are generally unknown in practice, approximations are used, which utilize observations made along the trajectory so far.
In the case of $g_k$, this is achieved by averaging the observations
\[
\widetilde g_k = \norm{f_{\omega_k}'(x_k)}^2
\]
using $p$-EMA to obtain an approximation $\widehat g_k$ to $g_k$.
This is motivated by the fact that $g_k$, which is an expectation, might be approximated by averaging over observations.
Clearly, this induces a biased estimator, as $x_k$ changes with $k$.
For the case of $\sigma_k$ one can't employ the same strategy, as this would require knowledge of $F'(x_k)$.
As a remedy, it is observed that (see \cite[Section 4.3]{KoehneKreisSchielaHerzog:2023:1})
\[
\sigma_k = \frac{\E{\xi_k, \xi_{}}{f_{\xi_{}}(x_{k+1}) - f_{\xi_k}(x_{k+1})}}{\alpha_k} + O(\alpha_k),
\]
where $\alpha_k$ is the step size used in iteration $k$.
Similarly as above, this motivates to use the observations
\[
\widetilde \sigma_k = \frac{f_{\xi_{k+1}}(x_{k+1}) - f_{\xi_k}(x_{k+1})}{\alpha_k}
\]
for averaging with $p$-EMA to obtain an approximation $\widehat \sigma_k$ to $\sigma_k$.
Using our results on the convergence of $p$-EMA we are able to describe the long-term behavior or the estimators obtained in this way.
This has further implications on the analysis of adaptive step size schemes, which are build upon these estimators, as well as on schemes that aim to detect stagnation of the algorithm, which happens in the non-interpolating setting (see \cref{subsec:interpolating-vs-non-interpolating} below).
One such consequence is the convergence of the estimated step sizes from \cite{KoehneKreisSchielaHerzog:2023:1} in the important non-interpolating setting (see \cref{subsec:interpolating-vs-non-interpolating}).
In the following, we will first recall the adaptive step sizes from \cite{KoehneKreisSchielaHerzog:2023:1} and subsequently present convergence results for the estimated step sizes.
\subsection{Adaptive Step Sizes}
In \cite{KoehneKreisSchielaHerzog:2023:1}, a theoretic adaptive step size rule leading to optimal convergence rates of SGD is identified as
\begin{equation}
\alpha_k = \frac{1}{L}\paren[auto](){1 - \frac{\Var{\xi}{\nabla f_{\xi}(x_k)}}{\E{\xi}{\norm{\nabla f_{\xi}(x_k)}^2}}}.
\label{eq:perfect-step-sizes}
\end{equation}
The goal is to use this step size in the $k$-th iteration of SGD.
Here the variance in the search direction is defined as:
\[
\Var{\xi}{\nabla f(\xi)} = \E{\xi}{\norm{\nabla f_\xi(x) - \nabla F(x)}^2}.
\]
Clearly, the step size rule \eqref{eq:perfect-step-sizes} is not computable exactly in practical scenarios, as the involved quantities are generally unknown.
As a remedy, the averaged quantities $\widehat g_k$ and $\widehat \sigma_k$ form above are used to approximate the quantities $g_k = \E{\xi}{\norm{\nabla f_\xi (x_k)}^2}$ and $\Var{\xi}{\nabla f_\xi(x_k)}$, respectively.
This leads to the \emph{practical} step size
\begin{equation}
\alpha_k = \frac{1}{L}\left(1 - \frac{\widehat \sigma_k}{\widehat g_k}\right)
\label{eq:practical-step-sizes}
\end{equation}
for the $k$-th iteration of SGD.
In the discussion in \cref{subsec:interpolating-vs-non-interpolating} we have stated that, in the non-interpolating setting, the step sizes $\alpha_k$ need to converge to zero to ensure convergence of the SGD algorithm.
Considering the practical step sizes \eqref{eq:practical-step-sizes}, such a convergence can only occur of
\begin{equation}
\zeta_k \coloneqq 1 - \frac{\widehat \sigma_k}{\widehat g_k}\to 0.
\label{eq:def-zeta}
\end{equation}
For the purpose of this presentation, we will assume that the factor $\frac{1}{L}$ is either known or can be approximated reasonably well, e.g. by line search methods.
Thus, we will focus on the term $\zeta_k$, as defined in \eqref{eq:def-zeta}.
\subsection{Exploiting Convergence of
\texorpdfstring{$p$-EMA}{p-EMA}
}
\label{subsec:exploiting-convergence}
Here, we elaborate the convergence of the estimators in simplified setting.
Consider SGD run with a sufficiently small, but constant step size $\alpha_k = \alpha$.
Evaluate the estimators $\widehat \sigma_k$ and $\widehat g_k$ as described above, but \textbf{not} use the suggested step sizes \eqref{eq:practical-step-sizes}.
Assume that SGD is sufficiently mixing.\footnote{In fact, it can be shown that under certain assumptions, the autocorrelations of the observables discussed below decay at a linear, thus summable, rate.}
Then, the existence of an invariant measure and the convergence results for $p$-EMA imply that almost surely:
\begin{equation}
\widehat g_k \to \int_\Omega \int_{\R^d}\norm{\nabla f_{\xi}(x)}^2 \intd \mu_\alpha^*(x)\intd P(\xi) =: g.
\label{eq:def-g-exp-value}
\end{equation}
Further, $\widehat \sigma_k$ converges almost surely to $\sigma$, given by:
\[
\sigma = \int_\Omega\int_{\R^d} \frac{F(x) - f_{\xi}(x - \alpha \nabla f_{\xi}(x))}{\alpha} \intd\mu_\alpha^*(x)\intd P(\xi).
\]
If $g = 0$, then $\mu_\alpha^*$ is a Dirac measure at the unique minimizer $x^*$ of $F$, which in this case is also a minimizer of every $f_\xi$, thus this is the case in the interpolating setting.
Here, it is known that we do not need $\alpha_k \to 0$ for convergence, but sufficiently small constant step sizes lead to linear rates of convergence of SGD.
Thus, from the perspective of step sizes estimation the non-interpolating setting is more interesting.
Here, if $g > 0$, we also have $\widehat \zeta_k \to 1 - \frac{\sigma}{g}$ almost surely.
We have by $L$-smoothness:
\[
f_{\xi}(x - \alpha \nabla f_{\xi}(x)) \le f_{\xi}(x) + \alpha\left(\frac{\alpha L}{2} - 1\right) \norm{\nabla f_{\xi}(x)}^2.
\]
Therefore:
\begin{align*}
\sigma
&
\ge
-\left(\frac{\alpha L }{2} - 1\right) \int_\Omega \int_{\R^d}\norm{\nabla f_{\xi}(x)}^2 \intd \mu_\alpha^*(x)\intd P(\xi)
\\
&
=
-\left(\frac{\alpha L }{2} - 1\right) g.
\end{align*}
Thus, for any $\eps > 0$, eventually,
\[
\zeta_k -\eps \le 1 - \frac{\sigma}{g} \le 1 - \frac{-\left(\frac{\alpha L }{2} - 1\right)g}{g} = \frac{\alpha L }{2}
\]
Therefore, the suggested step size $\widehat \alpha_k = \frac{1}{L} \zeta_k$ also converges almost surely to a limit $\alpha^* \le \frac{\alpha}{2}$.
Despite only applying to SGD with constant step sizes, this insight can open the door to a deeper understanding of SGD with the estimated step sizes $\widehat \alpha_k$, which one would use in a practical scenario.
The intuition behind this can be described as follows:
If, in the non-interpolating setting, for some reason, the step sizes might not decrease to zero (assume that $\widehat \zeta_k \in [0,1]$), but stagnate at some positive limit, this would prevent SGD from converging.
However, in this scenario, the results from above indicate that the step sizes will now eventually be reduced.
Thus, in the non-interpolating setting, the estimated step sizes cannot stagnate at any positive threshold and will decrease to zero, which in turn enables convergence of SGD.
Clearly, the above discussion is heuristic.
It indicates further directions of research, which elaborate the connection between the invariant measure, the speed of convergence towards this measure, and the behavior of the estimated step sizes.
As this paper dedicated to the development and analysis of $p$-EMA, a further discussion would exceed the intended scope of this motivating example.
\section{Numerical Studies}
\label{sec:numerics}
To illustrate the benefits of $p$-EMA and the findings of this paper, we perform a series of numerical experiments.
In a first experiment, presented in \cref{subsec:numerics-convergence}, we show how $p$-EMA achieves the properties described in the introduction (faster convergence on eventually stationary data, and convergence on stationary data, where EMA fails to converge).
In a second series of experiments, presented in \cref{subsec:numerics-sgd}, we compare $p$-EMA with EMA and the arithmetic mean as averaging techniques used for averaging the estimators $\widehat g_k$ and $\widehat \sigma_k$.
Here, we will demonstrate the convergence of the suggested step sizes below a threshold $\frac{\alpha}{2}$, when SGD is run with a constant step size $\alpha$, as described in \cref{subsec:exploiting-convergence}.
\subsection{Convergence Properties of
\texorpdfstring{$p$-EMA}{p-EMA}
}
\label{subsec:numerics-convergence}
With the first experiments we will demonstrate how $p$-EMA achieves the properties, which where described in the introduction.
In particular, we want to demonstrate that
\begin{enumerate}[label=\arabic*.]
\item If the mean of the data is not constant, but changes over time, $p$-EMA is more capable of following the trend than the classical arithmetic mean \eqref{eq:arithmetic-mean}.
\item If the data becomes stationary eventually, $p$-EMA is able to converge to the mean of the data (unlike classical EMA \eqref{eq:ema}).
\end{enumerate}
\begin{figure}
\includegraphics[width=\textwidth]{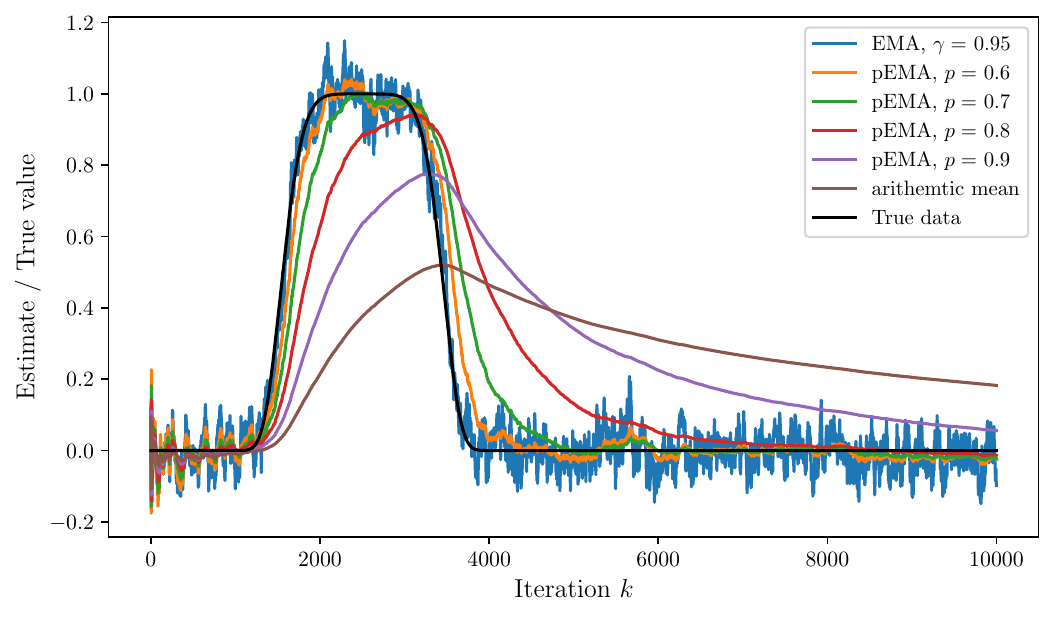}
\caption{Comparison of different averaging schemes on asymptotically stationary data.}
\label{fig:comparision:1}
\end{figure}
To this end, consider \cref{fig:comparision:1}.
In the experiment depicted there, we applied $p$-EMA (with different values of $p$ in the admissible range $p \in (\frac12, 1)$), classical EMA, and the arithmetic mean (which is identical to $p$-EMA with $p=1$) to noisy observations.
The noisy observations were made along the black curve, where an additive error was added independently to each of the 10,000 true observations.
As expected, EMA (blue curve) is the best averaging scheme to follow the true trajectory, however, it fails to converge to the mean once the process becomes stationary at approximately iteration 4,000.
As described in the introduction, this is due to the fact, that the weight assigned to the last observation is not decreasing to zero.
On the other hand, the arithmetic mean (brown curve) is able to converge to the mean of the stationary distribution, but it suffers from the observations made along the way, as all observations are assigned the same weight.
In contrast, $p$-EMA is, dependent on the parameter $p$, able to follow the curve of the true data quite well, and, additionally the noise in the estimator provided is reduced to zero over time, as it is expected.
Additional information on this experiment can be found in \cref{app:subsec:details-experiments-convergence-pema}.
\begin{figure}
\includegraphics[width=\textwidth]{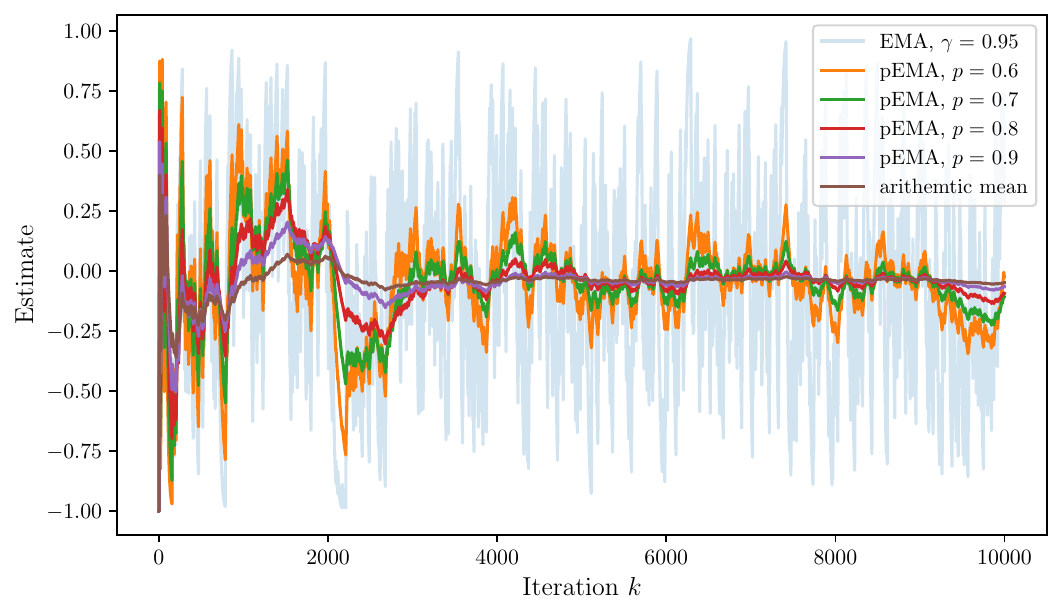}
\caption{Comparison of different averaging schemes on data generated by a jump process.}
\label{fig:comparision:2}
\end{figure}
In \cref{fig:comparision:2} we show the behavior of the averaging techniques applied to iteratively generated, stochastically depended data generated as follows:
Choose $x_1 \in \{-1,1\}$, then, iteratively:
\[
x_{k+1}
=
\begin{cases}
x_k, & \text{ with probability } q \\
-x_k, & \text{ with probability } 1 - q
\end{cases}
\]
with some fixed $q \in (0,1)$.
In our experiment we have used $q = \frac{9}{10}$.
This generates a stationary stochastic process with mean $0$ and invariant distribution $\pi = \mathcal U(\{-1,1\})$ being the uniform distribution on $\{-1,1\}$.
Again, EMA (here plotted transparently for better visibility of the other curves) fails to converge to the mean, due to the absence of noise reduction.
Also, as one would expect at a stationary process, the classic arithmetic mean converges and is the fastest averaging scheme to reduce the noise.
However, $p$-EMA also reduces the noise in its estimations, also leading to convergence to the mean.
\subsection{Adaptive SGD}
\label{subsec:numerics-sgd}
\begin{figure}
\includegraphics[width=\textwidth]{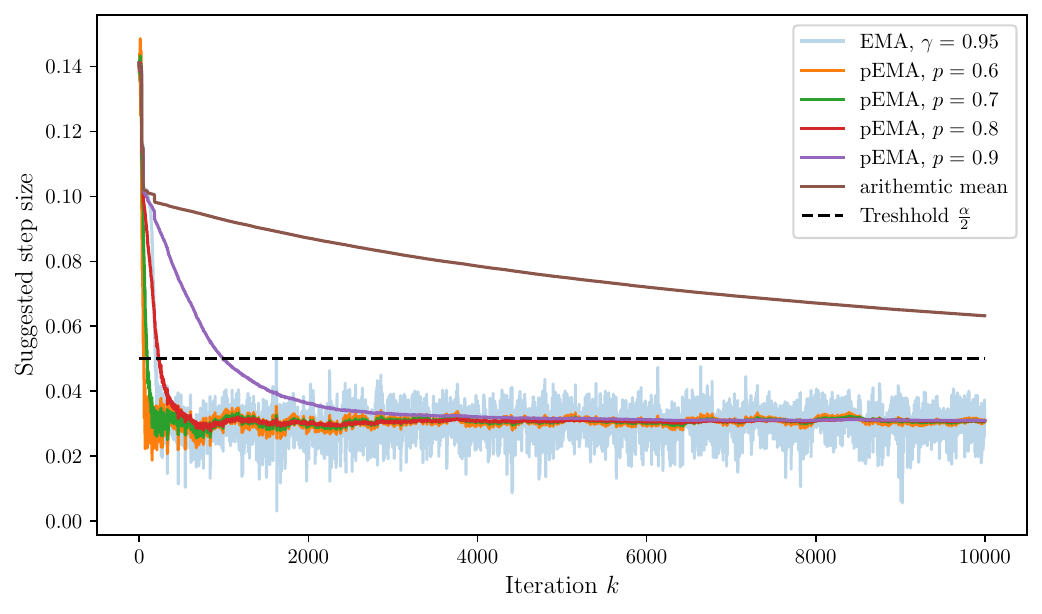}
\caption{Convergence of suggested step size: Artificial Problem.}
\label{fig:sugg-lrs-artificial}
\end{figure}

In this subsection, we will demonstrate the effect of employing $p$-EMA to average the observations used in the estimators as described in \cref{sec:implications}, and compare it to
the other averaging techniques discussed in this work, i.e. classical arithmetic mean \eqref{eq:arithmetic-mean} and classical EMA \eqref{eq:ema}.
We give numerical evidence for the convergence behavior described in \cref{subsec:exploiting-convergence}, namely that the \emph{suggested} step sizes $\frac{1}{L}\zeta_k$, where $\zeta_k$ is defined in \eqref{eq:def-zeta}, will converge to a value $\le \frac\alpha2$, when SGD is run with sufficiently small constant step size $\alpha$ and the estimation is made along the trajectory of SGD.
For this experiment we use a stochastic optimization problem, which fits the theoretical assumptions.
Details about how the problem is constructed can be found in \cref{app:subsec:details-quadratic-sop}.
\cref{fig:sugg-lrs-artificial} shows the development of the suggested step sizes, if SGD is run with a constant step size, but the estimators are computed as described in \cref{sec:implications} with the respective averaging techniques.
On the one hand, one can see that all averaging techniques reach the threshold $\frac{\alpha}{2}$ described in \cref{subsec:exploiting-convergence}.
On the other hand, one can see the different speed of convergence: The larger the $p$ in $p$-EMA, the more the old observations from the initialization, where SGD wasn't yet stationary, corrupt the estimation process in later stages.
Again, classical EMA also reaches the \emph{correct} mean fast, but fails to reduce the noise to produce a reliable estimate.
In fact, some estimates violate the threshold, even after the EMA estimate has apparently stabilized (see \cref{fig:sugg-lrs-artificial} at iteration approximately 1700).
\section{Conclusion}
We have proposed and analyzed a novel averaging technique, which is particularly suited for situations, where observations are made along trajectories of systems, which become stationary, but it is unknown \emph{when} the transition to a stationary distribution is happening.
In such scenarios, the estimation given by the classical arithmetic mean suffers from outdated observations, while classical EMA fails to converge.
$p$-EMA finds a careful balance between these two extremes, enhancing the ability to adapt to changes in the underlying distribution of the data, while maintaining convergence guarantees.
In the context of stochastic optimization, we have demonstrated, how $p$-EMA provides reliable estimates for quantities necessary for the construction of adaptive step size algorithms.
More generally, $p$-EMA can be applied to other averaging processes along trajectories of stochastic optimization algorithms, e.g. momentum updates, and our strong convergence results
open the door to a deeper theoretical understanding of such methods.

\appendix
\section{Details on the Numerical Experiments}
\subsection{Experiments in \cref{subsec:numerics-convergence}}
\label{app:subsec:details-experiments-convergence-pema}
\begin{figure}
\subfigure{
\centering
\includegraphics[width=.8\textwidth]{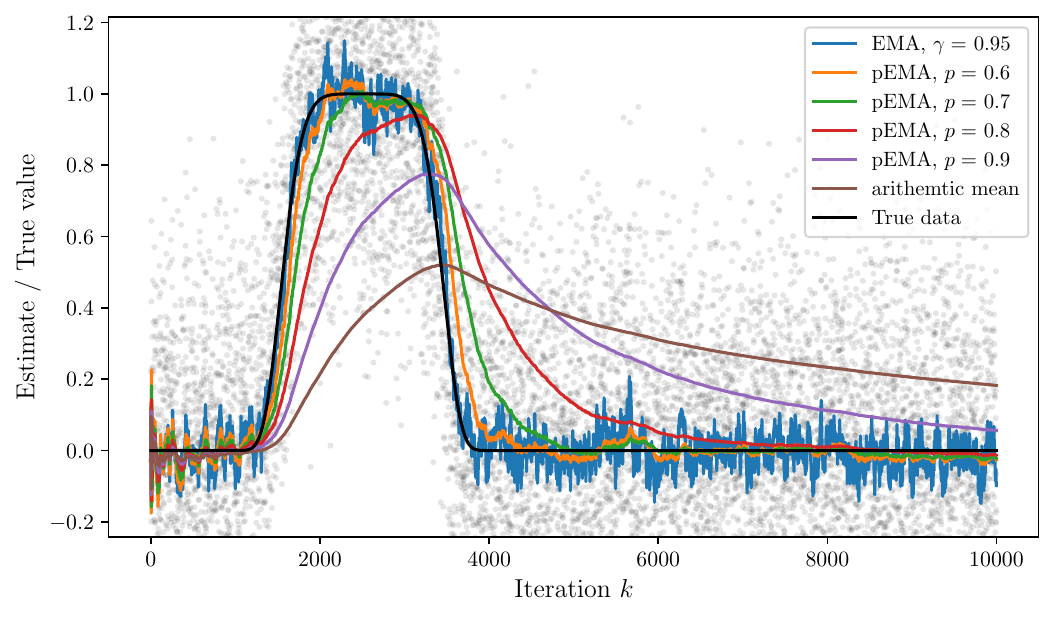}
}
~\subfigure{
\centering
\includegraphics[width=.8\textwidth]{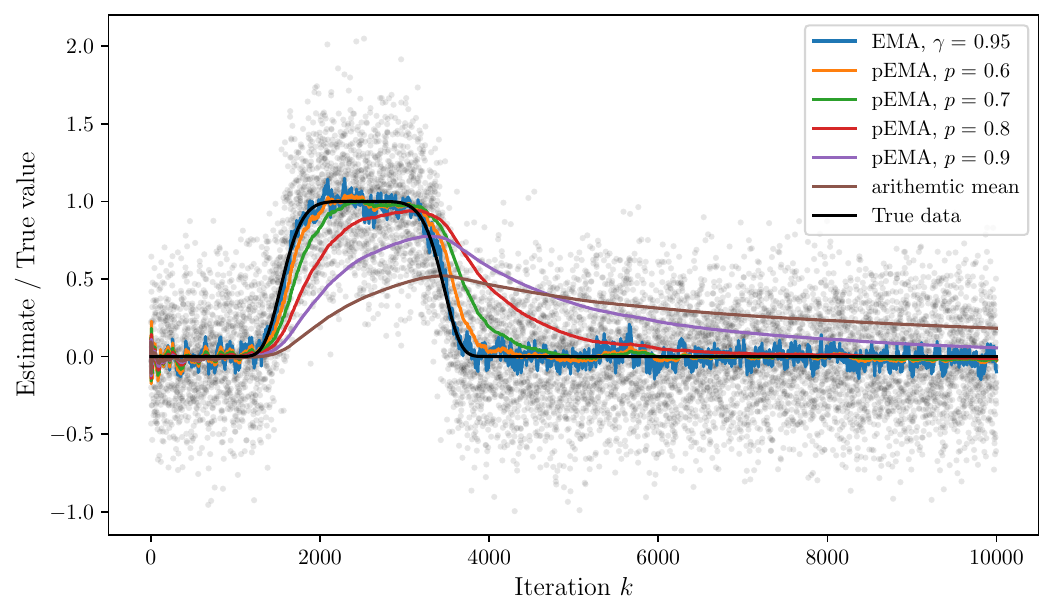}
}
\caption{Extended version of \cref{fig:comparision:1}, here with noisy observations plotted as black dots. Using the same vertical axis limits (left) and showing all noisy data points (right).}
\label{fig:comparision:1-extended}
\end{figure}

\begin{figure}
\subfigure{
\centering
\includegraphics[width=.8\textwidth]{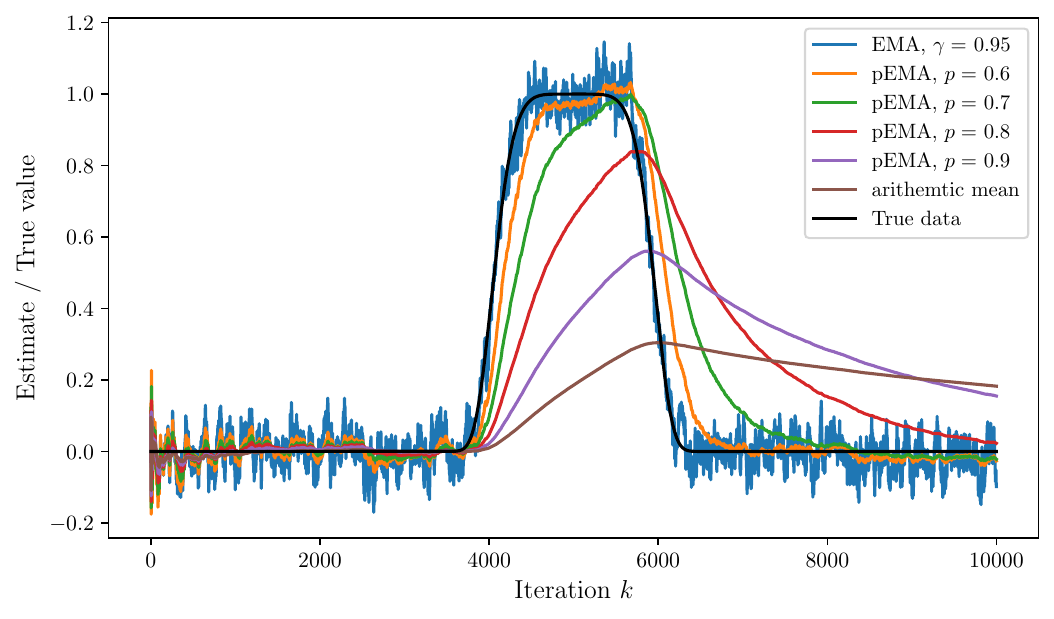}
}
\subfigure{
\centering
\includegraphics[width=.8\textwidth]{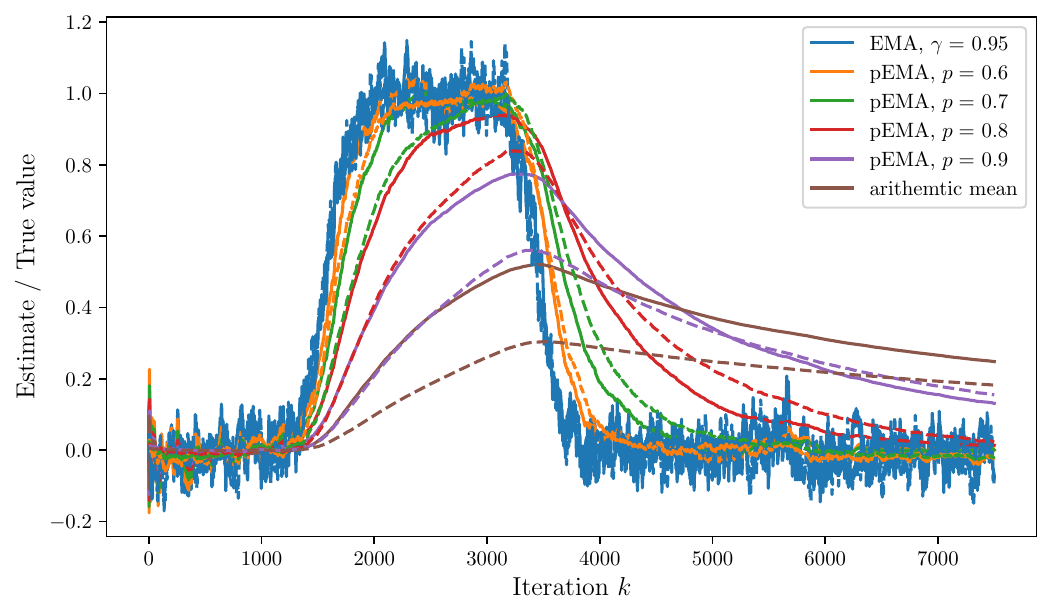}
}
\caption{Shifted version of experiment for \cref{fig:comparision:1}. Here the shift in distribution happens after a larger number of iterations. The plot on the left is analogous to \cref{fig:comparision:1}. On the right, we compare the behavior of the different averaging techniques with respect to their behavior during the distributional shift, by overlaying \cref{fig:comparision:1} and the figure on the left, and shifting such that the distributional shifts overlay. We plot the experiment from \cref{fig:comparision:1} with solid lines, and the experiment from the plot on the left of this figure with dashed lines.}
\label{fig:comparision:1-shift}
\end{figure}

In the first experiment (depicted in \cref{fig:comparision:1}), we considered observations made along a trajectory, which became stationary.
The black curve, labeled \emph{true data} is given here as evaluations of the function
\[
f(x) = \exp\left(-\left(\frac{x}{20}\right)^6\right)
\]
at an evenly spaced grid on $[-\frac{1}{2}, \frac{3}{2}]$.
To obtain noisy observations, normal distributed noise was added to each of the evaluations.
A version of \cref{fig:comparision:1}, which includes the noisy observations is given in \cref{fig:comparision:1-extended}.
An interesting question which arises in the study of different averaging techniques, is how sensitive the techniques are to a change in the underlying distribution.
As a general observation (for $p$-EMA and the arithmetic mean), this sensitivity decays, with number of iteration, where this change happens or starts.
We believe that this behavior is also crucial to obtain convergence: If this sensitivity was independent on the time, the distributional shift occurs, this would prevent almost convergence of the averaging technique, as it is the case in EMA.
An experiment which illustrates these thoughts is given in \cref{fig:comparision:1-shift}.
\subsection{Experiment in \cref{subsec:numerics-sgd}}
\label{app:subsec:details-quadratic-sop}
In the experiment depcited in \cref{fig:sugg-lrs-artificial}, we have compared the \emph{suggested} step sizes, which the different averaging techniques provide, along the same trajectory of constant step size SGD applied to a quadratic stochastic optimization problem.
The stochastic optimization problem we consider here is synthetic, and meets the assumptions of our theory on the convergence of the estimators.
We consider $f_\xi$ of the form
\[
f_\xi(x) = \frac12 x^T A_\xi x + b_\xi^T x,
\]
where $A_\xi$ and $b_\xi$ are constructed as follows:
We select a random orthogonal matrix $S \in \R^{d\times d}$ and a diagonal matrix $D = \diag(\lambda_1, \dots, \lambda_n)$.
We set the mean Hessian to $A \coloneqq S^\transp D S$ and select a noise level $\sigma_A > 0$.
In every iteration, we sample a random matrix $\Xi \in \R^{n \times n}$ with every entry $\xi_{ij}$ drawn from the uniform distribution on $[-\sigma_A,\sigma_A]$.
Then we let $W \coloneqq \Xi^\transp \Xi - \frac{2}{3} \sigma_A^3 \id$.
As is easily checked, this ensures $\E{\Xi}{W} = 0$.
We then use $A_\xi = A + W$ as the matrix for the quadratic SOP in the respective iteration.
For $b \in \R^d$, we choose a noise level $\sigma_b \ge 0$ and sample every entry of $b_\xi$ from the uniform distribution on $[-\sigma_b, \sigma_b]$.

\clearpage
\printbibliography

\end{document}